\documentclass{article} 
\usepackage{graphicx}
\usepackage{chicago}
\usepackage{times}

\evensidemargin	= \oddsidemargin
\newcommand{\appears}[4]{%
  \begin{picture}(0,0)
  \put(0,0){\raisebox{#1}{%
    \hspace*{#2}%
    \parbox[t]{#3}{\normalsize\tt #4}}}
  \end{picture}}

\def\paragraph#1{\bigskip\noindent\textbf{#1}}

\def\oab{\mathopen{<}}
\def\cab{\mathclose{>}}
\def\cw{\mathop{\oab c, w \cab}}
\def\class{{\cal C}}
\def\vocab{{\cal V}}
\def\event{{\cal E}}
\def\argmax{\mathop{\rm argmax}}

\def\org{$\oab$\textsf{\footnotesize organization}$\cab$}
\def\per{$\oab$\textsf{\footnotesize person}$\cab$}
\def\loc{$\oab$\textsf{\footnotesize location}$\cab$}
\def\dat{$\oab$\textsf{\footnotesize date}$\cab$}
\def\tim{$\oab$\textsf{\footnotesize time}$\cab$}
\def\mon{$\oab$\textsf{\footnotesize money}$\cab$}
\def\pct{$\oab$\textsf{\footnotesize percentage}$\cab$}
\def\oth{$\oab$\textsf{\footnotesize other}$\cab$}
\def\pls{$\oab$\textbf{\footnotesize +}$\cab$}

\def\corg{$\oab$\textsf{\footnotesize /organization}$\cab$}
\def\cper{$\oab$\textsf{\footnotesize /person}$\cab$}
\def\cloc{$\oab$\textsf{\footnotesize /location}$\cab$}
\def\cdat{$\oab$\textsf{\footnotesize /date}$\cab$}

\def\person#1{$\oab$\textsf{\footnotesize person,#1}$\cab$}

\def\dateclass#1{$\oab$\textsf{\footnotesize date,#1}$\cab$}

\def\other#1{$\oab$\textsf{\footnotesize other,#1}$\cab$}

\def\unk{\textsf{\small unknown}}
\def\muc7{\textsl{MUC--7}}
\def\hub4e{\textsl{Hub--4E}}
\def\iene{\textsl{IE--NE}}
\def\p{\emph{P}}
\def\r{\emph{R}}
\def\pr{\emph{P}\&\emph{R}}
\def\ser{\emph{SER}}
\def\wer{\emph{WER}}

\begin{document}

\title{Information Extraction \\ from Broadcast News}
\author{Yoshihiko Gotoh and Steve Renals \\ \\
	University~of~Sheffield, Department~of~Computer~Science \\
	Regent~Court, 211~Portobello~Street, Sheffield~~S1~4DP, UK \\
	\{y.gotoh,s.renals\}@dcs.shef.ac.uk}
\date{}
\maketitle
\appears{3.0in}{-0.2in}{5.0in}{
	Philosophical Transactions of the Royal Society of London, series A:
	Mathematical, Physical and Engineering Sciences, vol.~358, issue~1769,
	April 2000}

\begin{abstract}
  This paper discusses the development of trainable statistical models for
  extracting content from television and radio news broadcasts.
  In particular we concentrate on statistical finite state models for
  identifying proper names and other \emph{named entities} in broadcast speech.
  Two models are presented: the first represents name class information as
  a word attribute; the second represents both word-word and class-class
  transitions explicitly.
  A common $n$-gram based formulation is used for both models.
  The task of named entity identification is characterized by relatively
  sparse training data and issues related to smoothing are discussed.
  Experiments are reported using the DARPA/NIST \hub4e evaluation for North
  American Broadcast News.
\end{abstract}
\centerline{\bf keywords:
	named entity; information extraction; language modelling}

\section{Introduction} 
\label{sec: introduction}

Simple statistical models underlie many successful applications of speech and
language processing.
The most accurate document retrieval systems are based on unigram statistics.
The acoustic model of virtually all speech recognition systems is based on
stochastic finite state machines referred to as hidden Markov models (HMMs).
The language (word sequence) model of state-of-the-art large vocabulary speech
recognition systems uses an $n$-gram model ($[n-1]$th order Markov model),
where $n$ is typically 4 or less.
Two important features of these simple models are their trainability and
scalability: in the case of language modelling, model parameters are
frequently estimated from corpora containing up to $10^9$ words.
These approaches have been extensively investigated and optimized for speech
recognition, in particular, resulting in systems that can perform certain
tasks (e.g., large vocabulary dictation from a cooperative speaker) with
a high degree of accuracy.
More recently, similar statistical finite state models have been developed for
spoken language processing applications beyond direct transcription to enable,
for example, the production of structured transcriptions which may include
punctuation or content annotation.

In this paper we discuss the development of trainable statistical models for
extracting content from television and radio news broadcasts.
In particular, we concentrate on \emph{named entity} (NE) identification,
a task which is reviewed in \S\ref{sec: nereview}.
Section~\ref{sec: framework} outlines a general statistical framework for NE
identification, based on an $n$-gram model over words and classes.
We discuss two formulations of this basic approach.
The first (\S\ref{sec: form1}) represents class information as a word
attribute; the second (\S\ref{sec: form2}) explicitly represents word-word and
class-class transitions.
In both cases we discuss the implementation of the model and present results
using an evaluation framework based on North American broadcast news data.
Finally, in \S\ref{sec: discussion}, we discuss our work in the context of
other approaches to NE identification in spoken language and outline some
areas for future work.

\section{Named Entity Identification} 
\label{sec: nereview}

\subsection{Review} 

Proper names account for around 9\% of broadcast news output, and their
successful identification would be useful for structuring the output of
a speech recognizer (through punctuation, capitalization and tokenization),
and as an aid to other spoken language processing tasks, such as summarization
and database creation.
The task of NE identification involves identifying and classifying those words
or word sequences that may be classified as proper names, or as certain other
classes such as monetary expressions, dates and times.
This is not a straightforward problem.
While \textsf{Wednesday 1 September} is clearly a date, and
\textsf{Alan Turing} is a personal name, other strings, such as
\textsf{the day after tomorrow},
\textsf{South Yorkshire Beekeepers Association} and \textsf{Nobel Prize} are
more ambiguous.

NE identification was formalized for evaluation purposes as part of the 5th
Message Understanding Conference \shortcite{muc5:conf93}, and the evaluation
task definition has evolved since then.
In this paper we follow the task definition specified for the recent broadcast
news evaluation (referred to as \hub4e \iene) sponsored by DARPA and NIST
\cite{chinchor:manual98b}.
This specification defined seven classes of named entity: three types of
proper name (\loc, \per\ and \org) two types of temporal expression
(\dat\ and \tim) and two types of numerical expression (\mon\ and \pct).
According to this definition the following NE tags would be correct:
\begin{quote}
  \dat \textsf{Wednesday 1 September}\cdat \\
  \per \textsf{Alan Turing}\cper \\
  \textsf{the day after tomorrow} \\
  \org \textsf{South Yorkshire Beekeepers Association}\corg \\
  \textsf{Nobel Prize}
\end{quote}
\textsf{The day after tomorrow} is not tagged as a date, since only
``absolute'' time or date expressions are recognized; \textsf{Nobel} is not
tagged as a personal name, since it is part of a larger construct that refers
to the prize. Similarly, \textsf{South Yorkshire} is not tagged as a location
since it is part of a larger construct tagged as an organization.

Both rule-based and statistical approaches have been used for NE
identification.
\citeN{wakao:coling96} and \citeN{hobbs:inbook97} adopted grammar-based
approaches using specially constructed grammars, gazetteers of personal and
company names, and higher level approaches such as name co-reference.
Some grammar-based systems have utilized a trainable component, such as the
Alembic system \cite{aberdeen:muc95}.
The LTG system \cite{mikheev:muc98} employed probabilistic partial matching,
in addition to a non-probabilistic grammar and gazetteer look-up.

\citeN{bikel:anlp97} introduced a purely trainable system for NE
identification, which is discussed in greater detail in \citeN{bikel:ml99}.
This approach was based on an ergodic HMM (i.e., an HMM in which every state
is reachable from every state) where the hidden states corresponded to NE
classes, and the observed symbols corresponded to words.
Training was performed using an NE annotated corpus, so the state sequence was
known at training time.
Thus likelihood maximization could be accomplished directly without need for
the expectation-maximization (EM) algorithm.
The transition probabilities of this model were conditioned on both the
previous state and the previous word, and the emission probabilities attached
to each state could be regarded as a word-level bigram for the corresponding
NE class.

NE identification systems are evaluated using an unseen set of evaluation data:
the hypothesised NEs are compared with those annotated in a human-generated
reference transcription.\footnote{
  Inter-annotator agreement for reference transcriptions is around
  97--98\%~\cite{robinson:darpa99}.}
In this situation there are two possible types of error: \emph{type}, where an
item is tagged as the wrong kind of entity and \emph{extent}, where the wrong
number of word tokens are tagged.
For example,
\begin{center}
  \loc \textsf{South Yorkshire}\cloc\ \textsf{Beekeepers Association}
\end{center}
has errors of both type and extent since the ground truth for this excerpt is
\begin{center}
  \org \textsf{South Yorkshire Beekeepers Association}\corg \ .
\end{center}
These two error types each contribute $0.5$ to the overall error count, and
precision (\p) and recall (\r) can be calculated in the usual way.
A weighted harmonic mean (\pr), sometimes called the F-measure
\cite{vanrijsbergen:book79}, is often calculated as a single summary statistic:
\begin{displaymath}
  P\&R = \frac{2RP}{R+P} \ .
\end{displaymath}
In a recent evaluation, using newswire text, the best performing system
\shortcite{mikheev:muc98} returned a \pr\ of 0.93.
Although precision and recall are clearly informative measures,
\citeN{makhoul:darpa99} have criticized the use of \pr, since it implicitly
deweights missing and spurious identification errors compared with incorrect
identification errors.
They proposed an alternative measure, referred to as the slot error rate
(\ser), that weights three types of identification error equally.\footnote{
  \ser\ is analogous to word error rate (\wer), a performance measure for
  automatic speech transcription.
  It is obtained by $SER = (I+M+S)/(C+I+M)$ where $C$, $I$, $M$, and $S$
  denote the numbers of correct, incorrect, missing, and spurious
  identifications.
  Using this notation, precision and recall scores may be calculated as
  $R = C/(C+I+M)$ and $P = C/(C+I+S)$, respectively.}

\subsection{Identifying Named Entities in Speech} 

A straightforward approach to identifying named entities in speech is to
transcribe the speech automatically using a recognizer, then to apply
a text-based NE identification method to the transcription.
It is more difficult to identify NEs from automatically transcribed speech
compared with text, since speech recognition output is missing features that
may be exploited by ``hard-wired'' grammar rules or by attachment to
vocabulary items, such as punctuation, capitalization and numeric characters.

More importantly, no speech recognizer is perfect, and spoken language is
rather different from written language.
Although planned, low-noise speech (such as dictation, or a news bulletin read
from a script) can be recognized with a word error rate (\wer) of less than
10\%, speech which is conversational, in a noisy (or otherwise cluttered)
acoustic environment or from a different domain may suffer a \wer\ in excess
of 40\%.
Additionally, the natural unit seems to be the phrase, rather than the
sentence, and phenomena such as disfluencies, corrections and repetitions are
common.
It could thus be argued that statistical approaches, that typically operate
with limited context and very little notion of grammatical constructs, are
more robust than grammar-based approaches.
\citeN{appelt:darpa99} oppose this argument, and have developed a finite-state
grammar-based approach for NE identification of broadcast news.
However, this relied on large, carefully constructed lexica and gazetteers,
and it is not clear how portable between domains this approach is.
Some further discussion of rule-based approaches follows in
\S\ref{sec: discussion}.

Spoken NE identification was first demonstrated by \citeN{kubala:darpa98}, who
applied the model of \shortciteN{bikel:ml99} to the output of a broadcast news
speech recognizer.
An important conclusion of that work --- supported by the experiments reported
here --- was that the error of an NE identifier degraded linearly with \wer,
with the largest errors due to missing and spuriously tagged names.
Since then several other researchers, including ourselves, have investigated
the problem within the \hub4e evaluation framework.

Evaluation of spoken NE identification is more complicated than for text,
since there will be speech recognition errors as well as NE identification
errors (i.e., the reference tags will not apply to the same word sequence as
the hypothesised tags).
This requires a word level alignment of the two word sequences, which may be
achieved using a phonetic alignment algorithm developed for the evaluation of
speech recognizers \cite{fisher:icassp93}.
Once an alignment is obtained, the evaluation procedure outlined above may be
employed, with the addition of a third error type, \emph{content}, caused by
speech recognition errors.
The same statistics (\pr\ and \ser) can still be used, with the three error
types contributing equally to the error count.

\section{Statistical Framework} 
\label{sec: framework}

First, let $\vocab$ denote a vocabulary and $\class$ be a set of name classes.
We consider that $\vocab$ is similar to a vocabulary for conventional speech
recognition systems (i.e., typically containing tens of thousands of words,
and no case information or other characteristics).
In what follows, $\class$ contains the proper names, temporal and number
expressions used in the \hub4e \iene\ evaluation described above.
When there is no ambiguity, these named entities are referred to as
``name(s)''.
As a convention here, a class \oth\ is included in $\class$ for those words
not belonging to any of the specified names.
Because each name may consist of one word or a sequence of words, we also
include a marker \pls\ in $\class$, implying that the corresponding word is
a part of the same name as the previous word.
The following example is taken from a human-generated reference transcription
for the 1997 \hub4e Broadcast News evaluation data:
\begin{center}
  \textsf{\footnotesize AT THE}
	$\underbrace{\mbox{\textsf{\footnotesize
	RONALD REAGAN CENTER}}}_{\mbox{\org}}$
	\textsf{\footnotesize IN}
	$\underbrace{\mbox{\textsf{\footnotesize SIMI VALLEY}}}_{\mbox{\loc}}$
	$\underbrace{\mbox{\textsf{\footnotesize CALIFORNIA}}}_{\mbox{\loc}}$
\end{center}
The corresponding class sequence is
\begin{center}
  \oth\ \pls\ \org\ \pls\ \pls\ \oth\ \loc\ \pls\ \loc
\end{center}
because \textsf{\footnotesize SIMI VALLEY} and
\textsf{\footnotesize CALIFORNIA} are considered two different names by the
specification \shortcite{chinchor:manual98b}.

\begin{figure}
\centerline{\includegraphics[width=4.5in]{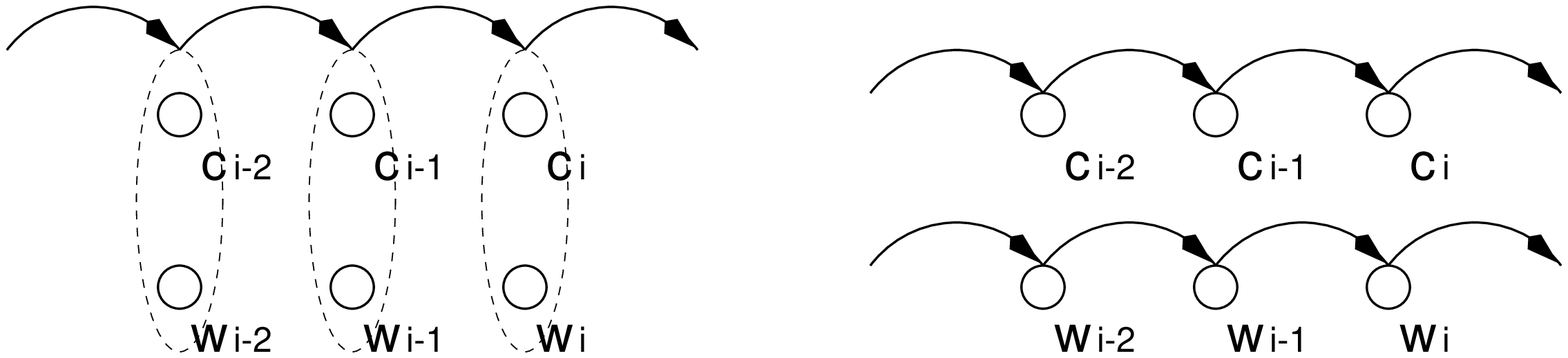}}
\caption[Topologies for NE models]
	{Topologies for NE models.
	The left model assumes that class information is a word attribute.
	The right model explicitly models word-word and class-class
	transitions.}
\label{fig: ne_model}
\end{figure}

Class information may be interpreted as a word attribute (the left model of
figure~\ref{fig: ne_model}).
Formally, we define a class-word token $\cw\in\class\times\vocab$ and consider
a probability
\begin{eqnarray}
  {\textstyle p(\cw_1, \cdots, \cw_m)}
	= \prod_{i=1 \cdots m} {\textstyle p(\cw_i | \cw_1, \cdots, \cw_{i-1})}
\label{eq: form1}
\end{eqnarray}
that generates a sequence of class-word tokens $\cw_1, \cdots, \cw_m$.
Alternatively, word-word and class-class transitions may be explicitly
formulated (the right model of figure~\ref{fig: ne_model}).
Then we consider a probability
\begin{eqnarray}
  p(c_1, w_1, \cdots, c_m, w_m)
	= \prod_{i=1 \cdots m} p(c_i, w_i | c_1, w_1, \cdots, c_{i-1}, w_{i-1})
\label{eq: form2}
\end{eqnarray}
that generates a sequences of words $w_1, \cdots, w_m$ and a corresponding
sequence of classes $c_1, \cdots, c_m$.
The first approach is simple and analogous to conventional $n$-gram language
modelling, however the performance is sub-optimal in comparison to the second
approach, which is more complex and needs greater attention to the smoothing
procedure.

For both formulations, we have performed experiments using data produced for
the \hub4e\ \iene\ evaluation.
The training data for this evaluation consisted of manually annotated
transcripts of the \hub4e Broadcast News acoustic training data (broadcast in
1996--97).
This data contained approximately one million words (corresponding to about
140 hours of audio).
Development was performed using the 1997 evaluation data (3 hours of audio
broadcast in 1996, about 32,000 words) and evaluation results reported on the
1998 evaluation data (3 hours of audio broadcast in 1996 and 1998, about
33,000 words).

\section{Modelling Class Information as a Word Attribute} 
\label{sec: form1}

In this section, we describe an NE model based on direct word-word transitions,
with class information treated as a word attribute.
This approach suffers seriously from data sparsity.
We briefly summarize why this is so.

\subsection{Technical Description} 

Formulation~(\ref{eq: form1}) may be best viewed as a straightforward
extension to standard $n$-gram language modelling.
Denoting $e=\cw$, (\ref{eq: form1}) is rewritten as
\begin{equation}
  {\textstyle p(e_1, \cdots, e_m)}
	= \prod_{i=1 \cdots m} {\textstyle p(e_i | e_1, \cdots, e_{i-1})}
\end{equation}
and this is identical to the $n$-gram model widely used for large vocabulary
speech recognition systems.
Because each token $e\in\class\times\vocab$ is treated independently, those
having the same word but the different class (e.g., \dateclass{MAY},
\person{MAY}, and \other{MAY}) are considered different members.
Using this formulation, class-class transitions are implicit.
Further it may be interpreted as a classical HMM, in which tokens $e_i$
correspond to states, with observations $c_i$ and $w_i$ generated from each
$e_i$.
Maximum likelihood estimates for model parameters can be obtained from the
frequency count of each $n$-gram given text data annotated with name
information.
Since the state sequence is known the forward-backward algorithm is not
required.
Standard discounting and smoothing techniques may be applied.

The search process is based on $n$-gram relations.
Given a sequence of words, $w_1, \cdots, w_m$, the most probable sequence of
names may be identified by tracing the Viterbi path across the class-word
trellis such that
\begin{eqnarray}
  \oab\hat{c}_1, \cdots, \hat{c}_m\cab
	= \argmax_{c_1 \cdots c_m} {\textstyle p(\cw_1, \cdots, \cw_m)} \ .
\label{eq: search1}
\end{eqnarray}
This process may be slightly elaborated by looking into a separate list of
names that augments $n$-grams of $\cw$ tokens.
Further technical details of this formulation are in \citeN{gotoh:esca99}.

\subsection{Experiment} 

Using the experimental setup described in \S\ref{sec: framework}, we estimated
a back-off trigram language model that contained $18,964$ class-word tokens in
a trigram vocabulary, with a further $3,697$ words modelled as unigram
extensions.

A hand transcription (provided by NIST) and four speech recognizer outputs
(three distributed by NIST representing the range of systems that participated
in the 1998 broadcast news transcription evaluation, and our own system
\cite{robinson:spcom2000}) were automatically marked with NEs, then scored
against the human-generated reference transcription.
The results are summarized in table~\ref{tb: eval1}.
The combined \pr\ score was about 83\% for a hand transcription.
For recognizer outputs, the scores declined as \wer\ increased.
As noted by other researchers (e.g., \citeN{miller:darpa99}) a linear
relationship between the \wer\ and the NE identification scores is observed.

\begin{table}
\begin{center}
\begin{tabular}{|r|c|c|c|c|c|} \hline
				& \wer	& \ser	& \r	& \p	& \pr \\
\hline\hline
  hand transcription (NIST)	& .000	& .286	& .799	& .865	& .831 \\
  recognizer output (NIST 1)	& .135	& .394	& .738	& .797	& .766 \\
		    (NIST 2)	& .145	& .399	& .741	& .791	& .765 \\
		    (NIST 3)	& .283	& .563	& .618	& .713	& .662 \\
  recognizer output (own)	& .210	& .452	& .700	& .769	& .733 \\
\hline
\end{tabular}
\caption[NE identification scores on 1998 \hub4e evaluation data, using the NE
	model with implicit class transitions]
	{NE identification scores on 1998 \hub4e evaluation data, using the NE
	model with implicit class transitions.
	A hand transcription and three recognizer outputs were provided by
	NIST.
	The bottom row is by our own recognizer.
	\wer\ and \ser\ indicate word and slot error rates.
	\r, \p, and \pr\ denote recall, precision, and a combined
	precision\&recall scores, respectively.
	This table contains further improvement since our participation in the
	1998 \hub4e evaluation.
	In this experiment, we used transcripts of Broadcast News acoustic
	training data (1996--97) for NE model generation, but did not rely on
	external sources.}
\label{tb: eval1}
\end{center}
\end{table}


We have previously made an error analysis of this approach
\shortcite{gotoh:esca99}, where it was observed that most correctly marked
names were identified through bigram or trigram constraints around each name
(i.e., the name itself and words before/after that name).
When the NE model was forced to back-off to unigram statistics, names were
often missed (causing a decrease in recall) or occasionally a bigram of words
attributed with another class was preferred (a decrease in precision).
For example consider the phrase
\begin{center}
  \textsf{\footnotesize ... DIRECTOR ADRIAN LAJOUS SAYS ...},
\end{center}
taken from the 1997 evaluation data, where \textsf{\footnotesize LAJOUS} was
not found in the vocabulary.
The maximum likelihood decoding for this phrase was:
\begin{center}
  ... \other{DIRECTOR} \other{unknown} \other{unknown} \other{SAYS} ...
\end{center}
Unigram statistics for \person{ADRIAN} and \person{unknown} existed in the
model, however none of the trigrams or bigrams outperformed a bigram entry
\begin{displaymath}
  p(\mbox{\other{SAYS}}\ |\ \mbox{\other{unknown}}) \ .
\end{displaymath}
Further, \other{unknown} had higher unigram probability than \person{ADRIAN},
and no other trigram or bigram was able to recover this name.
(There was no unigram entry for \other{ADRIAN}.)
As a consequence, \textsf{\footnotesize ADRIAN LAJOUS} was not identified as
\per.

This is an example of a data sparsity problem that is observed in almost every
aspect of spoken language processing.
Although NE models cannot accommodate probability parameters for a complete
set of $n$-gram occurrences, a successful recovery of name expressions is
heavily dependent on the existence of higher order $n$-grams in the model.
The implicit class transition approach contributes adversely to the data
sparsity problem because it causes the set of possible tokens to increase in
size from $|\vocab|$ to $|\class\times\vocab|$.

\section{Explicit Modelling of Class and Word Transitions} 
\label{sec: form2}

In this section, an alternative formulation is presented that explicitly
models constraints at the class level, compensating for the fundamental
sparseness of $n$-gram tokens on a vocabulary set.
Recent work by \shortciteN{miller:darpa99} and \citeN{palmer:darpa99} has
indicated that such explicit modelling is a promising direction as $P\&R$
scores of up to 90\% for hand transcribed data have been achieved using an
ergodic HMM.
These formulations may be regarded as a two-level architecture, in which the
state transitions in the HMM represent transitions between classes (upper
level), and the output distributions from each state correspond to the
sequence of words within each class (lower level).

The formulation developed here is simpler because, rather than introducing
a two-level architecture, we describe a flat state machine that models the
probabilities of the current word and class conditioned on the previous word
and class (the right model of figure~\ref{fig: ne_model}).
We do not describe this formulation as an HMM, as the probabilities are
conditioned both on the previous word and on the previous class.
Only a bigram model is considered; however it outperforms the trigram
modelling of \S\ref{sec: form1}.

\subsection{Technical Description} 

Formulation~(\ref{eq: form2}) treats class and word tokens independently.
Using bigram level constraints, (\ref{eq: form2}) is reduced to
\begin{eqnarray}
  p(c_1, w_1, \cdots, c_m, w_m)
	= \prod_{i=1 \cdots m} p(c_i, w_i | c_{i-1}, w_{i-1}) \ .
\label{eq: form21}
\end{eqnarray}
The right side of (\ref{eq: form21}) may be decomposed as
\begin{eqnarray}
  p(c_i, w_i | c_{i-1}, w_{i-1})
	= p(w_i | c_i, c_{i-1}, w_{i-1}) \cdot p(c_i | c_{i-1}, w_{i-1}) \ .
\label{eq: form22}
\end{eqnarray}
The conditioned current word probability $p(w_i | c_i, c_{i-1}, w_{i-1})$ and
the current class probability $p(c_i | c_{i-1}, w_{i-1})$ are in the same form
as a conventional $n$-gram, hence may be estimated from annotated text data.

The amount of annotated text data available is orders of magnitude smaller
than the amount of text data typically used to estimate $n$-gram language
models for large vocabulary speech recognition.
Smoothing the maximum likelihood probability estimates is therefore essential
to avoid zero probabilities for events that were not observed in the training
data.
We have applied standard techniques in which more specific models are smoothed
with progressively less specific models.
The following smoothing path was chosen for the first term on the right side
of (\ref{eq: form22}):
\begin{displaymath}
  p(w_i | c_i, c_{i-1}, w_{i-1})
	\longrightarrow p(w_i | c_i, c_{i-1})
	\longrightarrow p(w_i | c_i)
	\longrightarrow p(w_i)
	\longrightarrow \frac{1}{|\cal W|} \ ,
\end{displaymath}
where ${|\cal W|}$ is the size of the possible vocabulary that includes both
observed and unobserved words from the training text data
(i.e., ${|\cal W|}$ is sufficiently greater than $|\vocab|$).
We preferred smoothing to $p(w_i | c_i, c_{i-1})$, rather than to
$p(w_i | c_i, w_{i-1})$, since we believed that the former would be better
estimated from the annotated training data.

Similarly, the smoothing path for the current class probability (the final
term in (\ref{eq: form22})) was:
\begin{displaymath}
  p(c_i | c_{i-1}, w_{i-1})
	\longrightarrow p(c_i | c_{i-1})
	\longrightarrow p(c_i) \ .
\end{displaymath}
This assumes that each class occurs sufficiently in training text data;
otherwise, further smoothing to some constant probability may be required.

Given the smoothing path, the current word probability may be computed using
an interpolation method based on that of \citeN{jelinek:proc80}:
\begin{eqnarray}
  p(w_i | c_i, c_{i-1}, w_{i-1})\hspace{-0.1in}
	&=& \hspace{-0.1in}
	\hat{f}(w_i | c_i, c_{i-1}, w_{i-1})
\nonumber \\
	&& \hspace{-0.1in}
	+\ \{1 - \alpha(c_i, c_{i-1}, w_{i-1})\} \cdot p(w_i | c_i, c_{i-1})
\label{eq: form23}
\end{eqnarray}
where $\hat{f}(w_i | c_i, c_{i-1}, w_{i-1})$ is a discounted relative
frequency and $\alpha(c_i, c_{i-1}, w_{i-1})$ is a non-zero probability
estimate (i.e., the probability that $\hat{f}(w_i | c_i, c_{i-1}, w_{i-1})$
exists in the model).

Alternatively, the back-off smoothing method of \citeN{katz:assp87} could be
applied:
\begin{eqnarray}
  p(w_i | c_i, c_{i-1}, w_{i-1})
	= \left\{\begin{array}{l}
		\hat{f}(w_i | c_i, c_{i-1}, w_{i-1}) \qquad \, \mbox{if }
		\event(c_i, w_i | c_{i-1}, w_{i-1}) \ \mbox{exists}, \\
		\beta(c_i, c_{i-1}, w_{i-1})\cdot p(w_i | c_i, c_{i-1}) \qquad
		\qquad \quad \mbox{otherwise}.
	  \end{array}\right.
\label{eq: form24}
\end{eqnarray}
In~(\ref{eq: form24}), $\beta(c_i, c_{i-1}, w_{i-1})$ is a back-off factor and
is calculated by
\begin{eqnarray}
  \beta(c_i, c_{i-1}, w_{i-1})
	= \displaystyle \frac{1 - \alpha(c_i, c_{i-1}, w_{i-1})}
	{\displaystyle 1 - \sum_{w_i\in\event(c_i, w_i | c_{i-1}, w_{i-1})}
	\hat{f}(w_i | c_i, c_{i-1})}
\label{eq: form25}
\end{eqnarray}
where $\event(c_i, w_i | c_{i-1}, w_{i-1})$ implies the event such that
current class $c_i$ and word $w_i$ occur after previous class $c_{i-1}$ and
word $w_{i-1}$.\footnote{
  The weaker models --- $p(w_i | c_i, c_{i-1})$, $p(w_i | c_i)$, and $p(w_i)$
  --- may be obtained in a way analogous to that used for
  $p(w_i | c_i, c_{i-1}, w_{i-1})$.
  The smoothing approach is similar for the conditioned current class
  probabilities, i.e., $p(c_i | c_{i-1}, w_{i-1})$, $p(c_i | c_{i-1})$, and
  $p(c_i)$.}
Discounted relative frequencies and non-zero probability estimates may be
obtained from training data using standard discounting techniques such as
Good-Turing, absolute discounting, or deleted interpolation.
Further discussion for discounting and smoothing approaches should be referred
to, e.g., \shortciteN{katz:assp87} or \citeN{ney:pami95}.

Given a sequence of words $w_1, \cdots, w_m$, named entities can be identified
by searching the Viterbi path such that
\begin{eqnarray}
  \oab\hat{c}_1\cdots \hat{c}_m\cab
	= \argmax_{c_1\cdots c_m} p(c_1, w_1, \cdots, c_m, w_m) \ .
\label{eq: search2}
\end{eqnarray}
Although the smoothing scheme should handle novel words well, the introduction
of conditional probabilities for \unk\ (which represents those words not
included in the vocabulary $\vocab$) may be used to model unknown words
directly.
In practice, this is achieved by setting a certain cutoff threshold when
estimating discounting probabilities.
Those words that occur less than this threshold are treated as \unk\ tokens.
This does not imply that smoothing is no longer needed, but that
conditional probabilities containing the \unk\ token may occasionally pick up
the context correctly without smoothing with weaker models.
The drawback is that some uncommon words are lost from the vocabulary.
Below we compare two NE models experimentally: one with \unk\ and fewer
vocabulary words and the other without \unk\ but with more vocabulary words.

\subsection{Experiment} 

Experiments were performed using the evaluation conditions described in
\S\ref{sec: framework}.
Two NE models (with explicit class transitions) were derived from transcripts
of the hand annotated Broadcast News acoustic training data.
One model contained no \unk\ token; there existed 27,280 different words in
the training data, all of which were accommodated in the vocabulary list.
Another model selected 17,560 words (from those occurring more than once in
the training data) as a vocabulary and the rest (those occurring exactly once
--- nearly 10,000 words) were replaced by the \unk\ token.

\begin{figure}
\centerline{\includegraphics[width=5.0in]{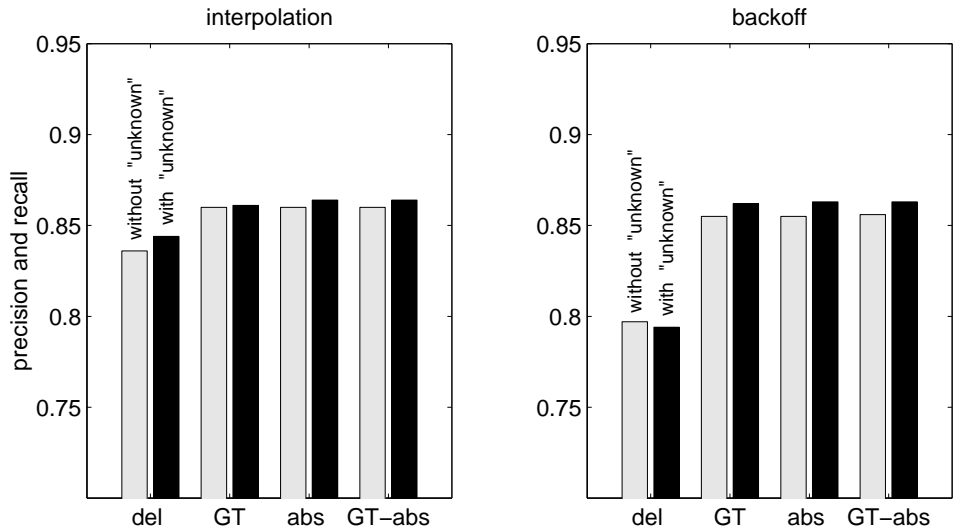}}
\caption[NE identification scores (\pr) on 1997 \hub4e hand transcription ---
	calculated using interpolation and back-off smoothing]
	{NE identification scores (\pr) on 1997 \hub4e hand transcription ---
	calculated using interpolation and back-off smoothing.
	NE models were built with and without the \unk\ token, using deleted
	interpolation (\textsf{\small del}), Good-Turing (\textsf{\small GT}),
	absolute (\textsf{\small abs}), and a combination of
	Good-Turing/absolute (\textsf{\small GT-abs}) discounting schemes.
	We have used 1997 data for a system development
	(as in figure~\ref{fig: p_and_r}), then applied to 1998 data for
	a system evaluation (as in table~\ref{tb: eval2}).}
\label{fig: p_and_r}
\end{figure}

Firstly, NE models were discounted using the deleted interpolation, absolute,
Good-Turing and combined Good-Turing/absolute discounting schemes.\footnote{
  The Good-Turing discounting formula is applied only when the inequality
  $rn_r \leq (r+1)n_{r+1}$ is satisfied, where $r$ is a sample count and $n_r$
  implies the number of samples that occurred exactly $r$ times.
  Empirically, and for most cases, this inequality holds only when $r$ is
  small.
  This may be modified slightly by applying absolute discounting to samples
  with higher $r$, which cannot be discounted using the Good-Turing formula
  (i.e., combined Good-Turing/absolute discounting).}
For each discounting scheme and with/without an \unk\ token,
figure~\ref{fig: p_and_r} shows \pr\ scores using the hand transcription of
the 1997 evaluation data.
For most cases, \pr\ was slightly better when \unk\ was introduced, although
the vocabulary size was substantially smaller.
Among discounting schemes, there was hardly any difference between Good-Turing,
absolute, and combined Good-Turing/absolute, regardless of the smoothing
method used.
Non-zero probability parameters derived using deleted interpolation did not
seem well matched to back-off smoothing.
We suspect, however, that the difference in performance would be negligible if
a sufficient amount of training data was available for the deleted
interpolation case.

Using \unk\ and the combined Good-Turing/absolute discounting scheme, followed
by back-off smoothing, table~\ref{tb: eval2} summarizes NE identification
scores for 1998 \hub4e evaluation data.
For the hand transcription and the four speech recognition outputs, this
explicit class transition NE model improved \pr\ scores by 4--6\% absolute
over the implicit model of \S\ref{sec: form1}.

\begin{table}
\begin{center}
\begin{tabular}{|r|c|c|c|c|c|} \hline
				& \wer	& \ser	& \r	& \p	& \pr \\
\hline\hline
  hand transcription (NIST)	& .000	& .187	& .863	& .922	& .892 \\
  recognizer output (NIST 1)	& .135	& .305	& .775	& .860	& .815 \\
		    (NIST 2)	& .145	& .296	& .779	& .867	& .821 \\
		    (NIST 3)	& .283	& .469	& .655	& .783	& .713 \\
  recognizer output (own)	& .210	& .381	& .729	& .823	& .773 \\
\hline
\end{tabular}
\caption[NE identification scores on 1998 \hub4e evaluation data, using the NE
	model with explicit class transitions]
	{NE identification scores on 1998 \hub4e evaluation data, using the NE
	model with explicit class transitions.
	A hand transcription and three recognizer outputs were provided by
	NIST.
	The bottom row is by our own recognizer.
	\wer\ and \ser\ indicate word and slot error rates.
	\r, \p, and \pr\ denote recall, precision, and a combined
	precision\&recall scores, respectively.
	The NE model contained 17,560 vocabulary words plus \unk\ token.
	A combination of Good-Turing/absolute discounting scheme was applied,
	followed by back-off smoothing.
	The best performing model in the 1998 \hub4e \iene\
	\shortcite{miller:darpa99} had \pr\ scores of .906, .815, .826, and
	.703 for the hand transcription and NIST recognizer outputs 1, 2, 3.}
\label{tb: eval2}
\end{center}
\end{table}


Although more complex in formulation, it is beneficial to model class-class
transitions explicitly.
Consider again the phrase
\textsf{\footnotesize ... DIRECTOR ADRIAN LAJOUS SAYS ...} discussed in
\S\ref{sec: form1}.
Here, \textsf{\footnotesize ADRIAN LAJOUS} was correctly identified as \per\
although \textsf{\footnotesize LAJOUS} was not included in the vocabulary.
It was identified using the product of conditional probabilities
\begin{displaymath}
  p(\mbox{\unk}\ |\ \mbox{\pls}, \mbox{\per})\cdot
	p(\mbox{\pls}\ |\ \mbox{\per}, \mbox{\textsf{\footnotesize ADRIAN}})
\end{displaymath}
between \textsf{\footnotesize ADRIAN} and \unk\, as well as the product
\begin{displaymath}
  p(\mbox{\textsf{\footnotesize SAYS}}\ |\
	\mbox{\oth}, \mbox{\per}, \mbox{\unk})\cdot
	p(\mbox{\oth}\ |\ \mbox{\per}, \mbox{\unk})
\end{displaymath}
between \unk\ and \textsf{\footnotesize SAYS}.

\subsection{An Alternative Decomposition} 

There exists an alternative approach to decomposing the right side of
Equation~(\ref{eq: form21}):
\begin{eqnarray}
  p(c_i, w_i | c_{i-1}, w_{i-1})
	= p(c_i | w_i, c_{i-1}, w_{i-1}) \cdot p(w_i | c_{i-1}, w_{i-1}) \ .
\label{eq: form26}
\end{eqnarray}
Theoretically, if the ``true'' conditional probability can be estimated,
decompositions by~(\ref{eq: form22}) and by~(\ref{eq: form26}) should produce
identical results.
This ideal case does not occur, and various discounting and smoothing
techniques will cause further differences between two decompositions.

In practice, the conditional probabilities on the right side
of~(\ref{eq: form26}) can be estimated in the same fashion as described in
\S\ref{sec: form1}: counting the occurrences of each token in annotated text
data, then applying certain discounting and smoothing techniques.
The adopted smoothing path for the current word probability was
\begin{displaymath}
  p(w_i | c_{i-1}, w_{i-1})
	\longrightarrow p(w_i | c_{i-1})
	\longrightarrow p(w_i)
	\longrightarrow \frac{1}{|\cal W|}
\end{displaymath}
and a path for the current class probability was
\begin{displaymath}
  p(c_i | w_i, c_{i-1})
	\longrightarrow p(c_i | w_i)
	\longrightarrow p(c_i) \ .
\end{displaymath}
In the latter case, a slight approximation
$p(c_i | w_i, c_{i-1}, w_{i-1}) \sim p(c_i | w_i, c_{i-1})$ was made, since it
was observed that $w_{i-1}$ did not contribute much when calculating the
probability of $c_i$ in this manner.

\begin{figure}
\centerline{\includegraphics[width=3.5in]{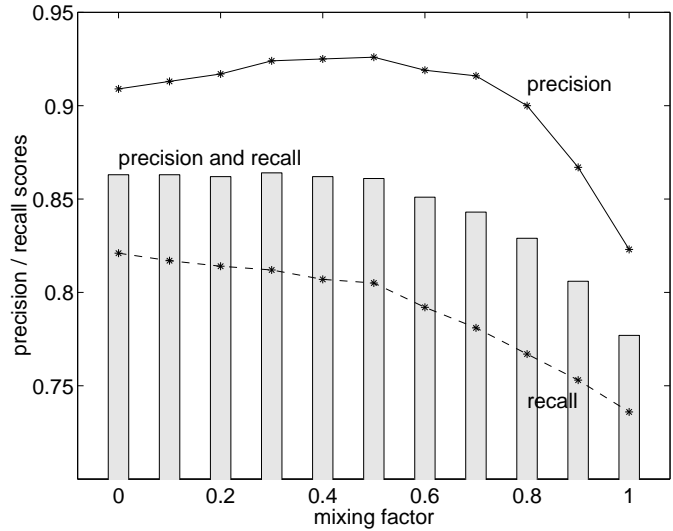}}
\caption[NE identification scores on 1997 hand transcription using mixtures of
	the two decompositions]
	{\pr\ scores on the 1997 hand transcription using mixtures of the two
	decompositions.
	NE models were built using \unk\, combined
	Good-Turing/absolute discounting, then back-off smoothing.}
\label{fig: dual}
\end{figure}

This second decomposition alone did not work as well as the initial
decomposition.
When applied to the 1997 hand transcription, the \pr\ score declined by 8\%
absolute (using \unk, combined Good-Turing/absolute discounting, and back-off
smoothing).
In general, decomposition by~(\ref{eq: form26}) accurately tagged words that
occurred frequently in the training data, but performed less well for uncommon
words.
Crudely speaking, it calculated the distribution over classes for each word;
consequently it had reduced accuracy for uncommon words with less reliable
probability estimates.
Decomposition by~(\ref{eq: form22}) makes a more balanced decision because it
relies on the distribution over words for each class, and there are orders of
magnitude fewer classes than words.

The two decompositions can be combined by
\begin{eqnarray}
  p(c_i, w_i | c_{i-1}, w_{i-1})
	= p_1(c_i, w_i | c_{i-1}, w_{i-1})^{1-k}
	\cdot p_2(c_i, w_i | c_{i-1}, w_{i-1})^k
\label{eq: form27}
\end{eqnarray}
where $p_1$ refers to the initial method and $p_2$ the alternative.
Figure~\ref{fig: dual} shows precision and recall scores for the mixture (with
factors $0.0\leq k\leq 1.0$) of the two decompositions.
It is observed that, for values of $k$ around $0.5$, this modelling improved
the precision without degrading the overall \pr.

\section{Discussion} 
\label{sec: discussion}

We have described trainable statistical models for the identification of named
entities in television and radio news broadcasts.
Two models were presented, both based on $n$-gram statistics.
The first model --- in which class information was implicitly modelled as
a word attribute --- was a straightforward extension of conventional language
modelling.
However, it suffered seriously from the problem of data sparsity, resulting in
a sub-optimal performance (a \pr\ score of 83\% on a hand transcription).
We addressed this problem in a second approach which explicitly modelled
class-class and word-word transitions.
With this approach the \pr\ score improved to 89\%.
These scores were based on a relatively small amount of training data
(one million words).
Like other language modelling problems, a simple way to improve the
performance is to increase the amount of training data.
\shortciteN{miller:darpa99} have noted that there is a log-linear relation
between the amount of training data and the NE identification performance; our
experiments indicate that the \pr\ score improves by a few percent for each
doubling of the training data size (between 0.1 and 1.0 million words).

The development of the second model was motivated by the success of the
approach of \shortciteN{bikel:ml99} and \shortciteN{miller:darpa99}.
This model shares the same principle of an explicit, statistical model of
class-class and word-word transitions, but the model formulation, and the
discounting and smoothing procedures differ.
In particular, the model presented here is a flat state machine, that is not
readily interpretable as a two-level HMM architecture.
Our experience indicates that an appropriate choice and implementation of
discounting/smoothing strategies is very important, since a more complex model
structure is being trained with less data, compared with conventional
language models for speech recognition systems.
The overall results that we have obtained are similar to those of
\shortciteANP{miller:darpa99}, but there are some differences which we cannot
immediately explain away.
In particular, although the combined \pr\ scores were similar,
\shortciteANP{miller:darpa99} reported balanced recall and precision, whereas
we have consistently observed substantially higher precision and lower recall.

The models presented here were trained using a corpus of about one million
words of text, manually annotated.
No gazetteers, carefully-tuned lexica or domain-specific rules were employed;
the brittleness of maximum likelihood estimation procedures when faced with
sparse training data was alleviated by automatic smoothing procedures.
Although the fact that an accurate NE model can be estimated from sparse
training data is of considerable interest and import, it is clear that it
would be of use to be able to incorporate much more information in
a statistical NE identifier.
To this end, we are investigating two basic approaches: the incorporation of
prior information; and unsupervised learning.

The most developed uses of prior information for NE identification are in the
form of the rule-based systems developed for the task.
Some initial work, carried out with Rob Gaizauskas and Mark Stevenson using
a development of the system described by \shortciteN{wakao:coling96}, has
analysed the errors of rule-based and statistical approaches.
This has indicated that there is a significant difference between the
annotations produced by the two systems for the three classes of proper name.
This leads us to believe that there is some scope for either merging the
outputs of the two systems, or incorporating some aspects of the rule-based
systems as prior knowledge in the statistical system.

Unsupervised learning of statistical NE models is attractive, since manual NE
annotation of transcriptions is a labour intensive process.
However, our preliminary experiments indicate that unsupervised training of NE
models is not straightforward.
Using a model built from 0.1 million words of manually annotated text, the
rest of the training data was automatically annotated, and the process
iterated.
\pr\ scores stayed at the same level (around 73\%) regardless of iteration.

Finally, we note that the NE annotation models discussed here --- and all
other state-of-the-art approaches --- act as a post-processor to a speech
recognizer.
Hence the strong correlation between the \pr\ scores of the NE tagger and the
\wer\ of the underlying speech recognizer is to be expected.
The development of NE models that incorporate acoustic information such as
prosody \cite{hakkanitur:euro_sp99} and confidence measures
\cite{palmer:euro_sp99} are future directions of interest.

\paragraph{Acknowledgements.}
  We have benefited greatly from cooperation and discussions with
  Robert Gaizauskas and Mark Stevenson.
  We thank BBN and MITRE for the provision of manually-annotated training data.
  The evaluation infrastructure was provided by MITRE, NIST and SAIC.
  This work was supported by EPSRC grant GR/M36717.

\bibliographystyle{chicago}

\label{lastpage}
\end{document}